\documentclass[times,review,10pt]{elsarticle}

\usepackage{amssymb}
\usepackage{amsmath}

\usepackage{graphicx}
\usepackage{float}
\usepackage{placeins} 

\usepackage{placeins}

\journal{Pattern Recognition}

\begin{document}

\begin{frontmatter}

\makeatother



\title{Quantum Vision Theory Applied to Audio Classification for Deepfake Speech Detection}



\author{Khalid Zaman}
\author{Melike Sah}
\author{Anuwat Chaiwongyen}
\author{Cem Direkoglu}







\begin{abstract}
We investigate the use of Quantum Vision (QV) theory, originally designed for image classification, for deep learning–based audio classification and demonstrate its effectiveness for deepfake speech detection. Inspired by particle–wave duality in quantum physics, QV theory is based on the idea that data can be represented not only in its observable, collapsed form, but also as information waves. In conventional deep learning, models are trained directly on these collapsed representations, such as images. In QV theory, inputs are first transformed into information waves using a QV block, and then fed into deep learning models for classification. QV-based models improve performance in image classification compared to their non-QV counterparts. What if QV theory is applied speech spectrograms for audio classification tasks? This is the motivation and novelty of the proposed approach. In this work, Short-Time Fourier Transform (STFT), Mel-spectrograms, and Mel-Frequency Cepstral Coefficients (MFCC) of speech signals are converted into information waves using the proposed QV block and used to train QV-based Convolutional Neural Networks (QV-CNN) and QV-based Vision Transformers (QV-ViT). Extensive experiments are conducted on the ASVSpoof dataset for deepfake speech classification. The results show that QV-CNN and QV-ViT consistently outperform standard CNN and ViT models, achieving higher classification accuracy and improved robustness in distinguishing genuine and spoofed speech. Moreover, the QV-CNN model using MFCC features achieves the best overall performance on the ASVspoof dataset, with an accuracy of  94.20\% and an EER of 9.04\%, while the QV-CNN with Mel-spectrograms attains the highest accuracy of 94.57\%. These findings demonstrate that QV theory is an effective and promising approach for audio deepfake detection and opens new directions for quantum-inspired learning in audio detection tasks.
\end{abstract}



\end{frontmatter}




\section{Introduction}
Audio classification is a fundamental task in speech and audio signal processing and plays an important role in applications such as automatic speech recognition, speaker verification, emotion recognition, and deepfake speech detection \cite{zaman2023survey}. In recent years, rapid advances in speech synthesis and voice conversion technologies have made artificially generated speech increasingly realistic \cite{shen2018natural, kim2021conditional, li2021starganv2}. This progress has raised serious concerns about security, privacy, and trust in voice-based systems, as deepfake speech threatens applications such as biometric authentication, financial services, and voice-driven interfaces \cite{todisco2019asvspoof, jung2025spoofceleb, chaiwongyen2026deepfake}. Consequently, developing robust techniques to detect manipulated or synthesized speech has become an important research problem.


Most existing research on audio classification relies on deep learning approaches that convert raw audio signals into time–frequency representations such as Short-Time Fourier Transform (STFT) \cite{costa2017evaluation, zaman2020classification}, Mel spectrograms \cite{luo2022vision, zhu2023multiscale}, and Mel-Frequency Cepstral Coefficients (MFCC) \cite{al2021rethinking}, which are treated as representations of 2D spectrogram images. These representations are then processed using deep learning architectures originally developed for computer vision, including Convolutional Neural Networks (CNNs) and Vision Transformers (ViTs). CNN-based models effectively capture local time–frequency patterns, while transformer-based models leverage self-attention mechanisms to model long-range dependencies and global context in spectrogram representations.

In general audio classification has been extensively studied using deep learning models and spectral feature representations. Spectrogram-based approaches, particularly those using FFT-derived or Mel-spectrogram inputs, have shown strong performance across environmental sound, speech, and music classification tasks. A spectrogram transformer architecture combining FFT and attention mechanisms was introduced for audio classification \cite{zhang2022spectrogram}. Patch-level Vision Transformers that convert audio signals into Mel-spectrogram patches and leverage ImageNet and AudioSet pre-training were also proposed \cite{luo2022vision}. Comparative studies further demonstrated the effectiveness of transformer models with transfer learning for spectrogram-based audio classification across multiple datasets \cite{nogueira2022transformers}. To improve efficiency for edge deployment, tiny transformer architectures inspired by BERT and trained on Mel-spectrogram images were developed\cite{wyatt2021environmental}. Self-supervised learning approaches, including masked spectrogram modeling, were introduced to enhance representation learning in low-resource scenarios \cite{gong2022ssast}. Patchout spectrogram transformers were proposed to improve training efficiency and generalization \cite{koutini2022efficient}. Swin Transformers were adapted for hierarchical spectrogram modeling for for music classification \cite{zhao2022s3t}. ASiT extended transformer-based spectrogram processing for audio tasks \cite{ahmed2024asit}. Causal Audio Transformers (CAT) were introduced for spectrogram modeling \cite{liu2023cat}. Multi-scale AST (MAST) variants explored hierarchical spectrogram representations \cite{chen2022hts}, \cite{zhu2023multiscale}. Convolution-free and multi-modal transformer frameworks further advanced spectrogram-based audio representation learning \cite{akbari2021vatt}.

In the domain of deepfake speech detection,the ASVspoof 2019 dataset has become a benchmark for deepfake speech detection. Early spoof detection systems based on CQCC-GMM and LFCC-GMM established baseline approaches using cepstral features \cite{nautsch2021asvspoof}. SE-ResNet, X-vector TDNN, DenseNet, MobileNetV2, ShuffleNetV2, and MNASNet were introduced as deep learning architectures operating on spectrogram and cepstral representations \cite{yoon2022bpcnn}. CNN-LSTM models were further explored by employing MFCC features for spoofed speech detection \cite{altalahin2023unmasking}. ResNet-34 was also applied for improved spoof detection performance \cite{aravind2020audio}. Ensemble models combining multiple architectures were proposed to enhance robustness \cite{chettri2019ensemble}. LCNN-based models were introduced using spectral and temporal modulation representations as input for spoofed speech detection \cite{cheng2023analysis}. Transformer-based architectures, including the Spectrogram Constant-Q Vision Transformer, extended attention-based modeling to spectrogram inputs \cite{ulutas2023deepfake}. Compact Convolutional Transformers (CCT) were also applied to deepfake detection tasks \cite{bartusiak2021synthesized}. Hybrid CNN- and ResNet-based frameworks such as Spec+ResNet+CE, Spec+SENet34, and Dilated ResNet were investigated for spectrogram-based spoof detection\cite{lai2019assert} . CQCC-DNN, eCQCC-DNN, and CQSPIC-DNN models explored cepstral feature learning with deep architectures \cite{das2020assessing}. Front-end modeling approaches such as IFCC, CQ-EST, CQ-OST, CQSPIC, and CMC-DNN were introduced to enhance feature extraction \cite{das2019long}. Spectrogram-CNN, MFCC-ResNet, Spec-ResNet, and CQCC-ResNet and fusion architectures were also explored for spoof detection \cite{nosek2019synthesized}, \cite{alzantot2019deep}. Spectrogram images with CNN and  Siamese CNN models using Gaussian probability feature were further investigated for deepfake detection \cite{lei2020siamese},  \cite{bartusiak2021frequency}.
Despite significant architectural advancements and strong development performance, evaluation results remain imperfect. Moreover, these approaches treat spectrograms as fixed, collapsed representations of speech signals, focusing primarily on classifier design rather than reconsidering the representation itself.
 
Therefore, in this work, we introduce a new perspective in deep learning, called Quantum Vision (QV) theory, and apply it to audio classification for deepfake speech detection. Inspired by the concept of particle–wave duality in quantum physics, QV theory is motivated by the idea that signals can be represented not only in their observable, collapsed form, but also as information wave functions that preserve richer characteristics of the data. In quantum physics, an unobserved object behaves as a wave that encodes all possible information and collapses into a particle only when it is observed or measured. Conventional deep learning models typically operate on collapsed representations such as spectrogram images, potentially losing useful information contained in the original signal.

QV theory proposes transforming conventional representations into information waves before feeding them into deep neural networks. This transformation is performed by a dedicated deep learning module called the QV block \cite{direkouglu2025quantum}. Instead of directly using spectrogram images for classification, the proposed approach converts them into quantum-inspired wave representations, which are then processed by deep learning models. The key question motivating this work is: \textit{Can transforming speech spectrograms into information waves improve audio classification, specifically in deepfake speech detection, compared to using standard spectrogram images?}

Motivated by this question, we apply QV theory to STFT and MFCC spectrograms for deepfake speech classification. The proposed QV block is integrated into Convolutional Neural Networks (QV-CNN) and Vision Transformers (QV-ViT), enabling end-to-end training. Extensive experiments are conducted on the ASVspoof dataset, a widely used benchmark for audio spoofing and deepfake detection. Experimental results demonstrate that QV-based models consistently outperform their non-QV counterparts, achieving higher classification accuracy and improved robustness in distinguishing genuine and spoofed speech signals.

The main contributions of this work are: We apply Quantum Vision (QV) theory to audio classification for deepfake speech detection and propose a QV block that converts speech spectrograms into quantum-inspired information wave representations. The proposed block is integrated into Convolutional Neural Networks (CNNs) and Vision Transformers (ViTs), forming QV-CNN and QV-ViT architectures. Extensive experiments using STFT and MFCC spectrograms are conducted on the ASVspoof dataset, and the results show that QV-based models consistently outperform standard deep learning models without QV.

The remainder of the paper is organized as follows. Section II reviews related work on deepfake speech detection and audio classification. Section III discusses the theoretical motivation behind QV theory. Section IV presents the proposed QV block and model architectures. Section V reports experimental results and comparisons. Finally, Section VI concludes the paper and outlines future research directions.

\section{Proposed Information-Wave Representation of Audio Spectrograms Using Quantum Vision Theory }

 \begin{figure*}[t]
\centering
\includegraphics[width=1\linewidth]{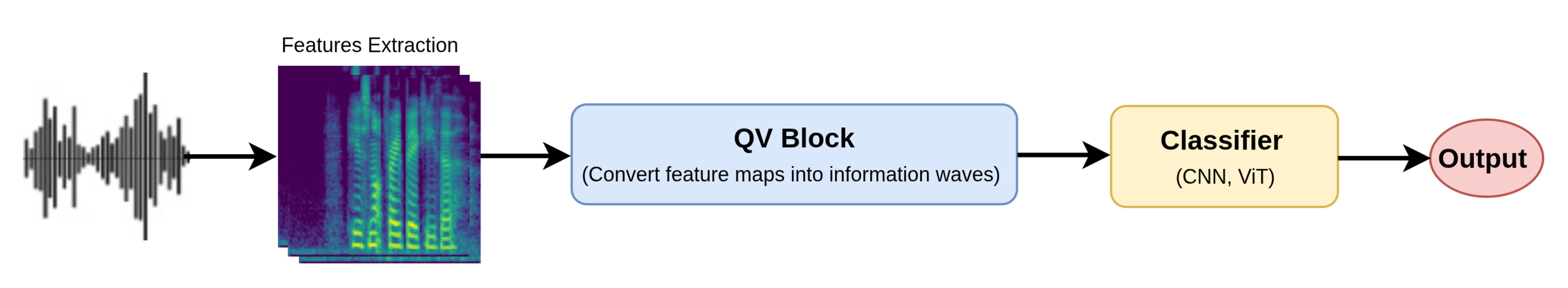} 
\caption{Block diagram of the propose study.}

\label{Fig:1}
\end{figure*}

Quantum Vision (QV) Theory is motivated by the behavior of quantum-scale objects, which exhibit particle-wave duality \cite{direkouglu2025quantum}. At the quantum level, particles such as photons and electrons behave as waves when unobserved, carrying all possible information about their position, momentum, and energy. When measured, these waves collapse into definite particle states. Schrödinger’s wave equation describes this wave behavior mathematically, while the uncertainty principle limits precise knowledge of a particle’s properties. QV Theory extends this concept to human-scale objects, suggesting that each object can be associated with an information wave function that contains all the possible information about it before perception. Observation, whether by a human or a measuring system, then collapses this wave into a perceivable object with definite characteristics.

In the human vision system, light reflected from objects enters the eye, passes through the cornea and lens, and forms an inverted image on the retina. The retina converts the light into electrical signals, which travel through biological neural networks to the Visual Cortex. During this process, past experiences and emotional states influence perception, so different observers may perceive the same scene differently. Within the QV framework, the incoming object information carried by light waves can be seen as a wave function that collapses in the Visual Cortex, allowing humans to recognize and locate objects. This provides a natural analogy between quantum measurement and human perception.

In machine vision, images are captured by cameras and processed directly by artificial neural networks, without the intermediate wave-like processing seen in human perception. QV Theory proposes a mathematical model that generates information wave functions for each object and feeds them into deep neural networks for classification. These wave functions can be described using quantum numbers, which encode discrete properties such as energy and momentum, capturing more detailed information about each object. By using object waves instead of fixed images, QV-inspired systems aim to bridge the gap between human visual perception and machine vision, potentially improving object recognition by leveraging wave-like representations.


We further extend the principles of QV Theory from visual object recognition to audio classification. In this approach, audio signals are first transformed into spectrogram images, which represent the frequency and amplitude content of sounds over time. These spectrograms are treated as “objects,” and the same QV-inspired wave function modeling is applied, generating information waves for each spectrogram. By feeding these audio wave functions into deep neural networks, the system can classify sounds in a manner analogous to object recognition in images. The overall framework of the proposed method is illustrated in Fig.~\ref{Fig:1}, suggesting that QV principles may provide a unified framework for processing information across multiple sensory modalities.

\subsection{QUANTUM VISION THEORY MATHEMATICAL MODEL}

In this section, we describe the construction of wave functions for spectrogram images using quantum principles. For a principal quantum number $n=3$, the corresponding quantum numbers $l=0,1,2$ and $m=0,\pm1,\pm2$ are used to model motion and direction; since an image is a two-dimensional function $I(x,y)$, the magnetic quantum number $m$ is applied along spatial directions to form basis wave functions. Specifically, in the $x$-direction, wave functions are generated by subtracting the original image from its shifted versions $I(x-m,y)$ for $m=\pm1,\pm2$, while $m=0$ produces no wave function because the subtraction yields zero, resulting in four basis wave functions as defined in Equation (1).

\begin{equation}
\begin{aligned}
\psi_{x,-1}(x,y) &= I(x+1,y)-I(x,y) \\
\psi_{x,+1}(x,y) &= I(x-1,y)-I(x,y) \\
\psi_{x,-2}(x,y) &= I(x+2,y)-I(x,y) \\
\psi_{x,+2}(x,y) &= I(x-2,y)-I(x,y)
\end{aligned}
\end{equation}

And for the $y$- direction,
\[
\psi_{y,m}(x,y)=I(x,y-m)-I(x,y)
\]
that creates four basis wave functions for $m=\pm1,\pm2$:

\begin{equation}
\begin{aligned}
\psi_{y,-1}(x,y) &= I(x,y+1)-I(x,y) \\
\psi_{y,+1}(x,y) &= I(x,y-1)-I(x,y) \\
\psi_{y,-2}(x,y) &= I(x,y+2)-I(x,y) \\
\psi_{y,+2}(x,y) &= I(x,y-2)-I(x,y)
\end{aligned}
\end{equation}

In total, eight basis wave functions are generated for $m=\pm1,\pm2$ across both spatial directions. Inspired by Schrödinger’s wave equation, where the probability distribution is given by $|\psi|^{2}$, subtracting the original image from its shifted versions and computing the magnitude square emphasizes object boundaries and exterior regions, thereby revealing spatial position information as shown in Equations (1) and (2).

According to quantum wave mechanics, the superposition of the basis wave functions defined in Equations (1) and (2) also forms a valid wave function that characterizes the object. Therefore, linear combinations of these basis wave functions, as expressed in Equation (3), can generate more informative representations.

\begin{equation}
\psi=\sum_{m=-2}^{2} a_m \psi_{x,m}(x,y)+b_m \psi_{y,m}(x,y)
\tag{3}
\end{equation}

where $a_m$ and $b_m$ are constant scalars, and $m=\pm1,\pm2$. Here, we have linear combination of 8 basis wave functions. Non-linear combination of these basis wave function are combined as shown below:

\begin{equation}
\begin{aligned}
\psi = \sum_{m=-2}^{2} \Big(
& \mathrm{ReLU}\!\left(H_m * \psi_{x,m}(x,y)\right) \\
+ \; &
\mathrm{ReLU}\!\left(V_m * \psi_{y,m}(x,y)\right)
\Big)
\end{aligned}
\tag{4}
\end{equation}

where $*$ is a convolution operation. $H_m$ and $V_m$ are convolution kernels for $m=\pm1,\pm2$. The convolution operation itself is a linear operator. ReLU is the linear rectifier unit that achieves the non-linear combination of basis functions. Here, we start to build a CNN that will find the best non-linear combination of basis wave functions to create informative wave functions describing spectrogram characteristics. We can also use colour (RGB) image as an original input image $I(x,y,C)$ where $C=[R,G,B]$ that generates basis wave functions as $\psi_{x,m}(x,y,C)$ and $\psi_{y,m}(x,y,C)$. Additionally, instead of using a single convolutional kernel within a CNN layer, multiple kernels (for example, 128) can be employed. After the nonlinear combination of the eight basis functions, this approach produces 128 distinct wave-function representations of the spectrogram, as illustrated in the equation below.

\begin{equation}
\begin{aligned}
\psi_{128} = \sum_{m=-2}^{2} \Big(
& \mathrm{ReLU}\!\left(H_{m,128} * \psi_{x,m}(x,y,C)\right) \\
+ \; &
\mathrm{ReLU}\!\left(V_{m,128} * \psi_{y,m}(x,y,C)\right)
\Big)
\end{aligned}
\tag{5}
\end{equation}

where $H_{m,128}$ and $V_{m,128}$ indicate that the convolutional layer has 128 kernels (filters) for $m=\pm1,\pm2$, and $\psi_{128}$ shows that 128 wave functions are created for a spectrogram image. Furthermore, each basis wave function can be processed through multiple convolutional layers followed by ReLU activations. The outputs of these layers are subsequently combined via summation to construct the spectrogram information waves. As an example, the formulation with three convolutional layers is presented below:

\begin{equation}
\begin{aligned}
\psi_{128} = \sum_{m=-2}^{2} \Big(
& \mathrm{ReLU}\!\Big(
H_{3,m,128} * \mathrm{ReLU}\!\big(
H_{2,m,128} *  \\
& \mathrm{ReLU}\!\big(
H_{1,m,128} * \psi_{x,m}(x,y,C)
\big)
\big)
\Big) \\
+ \; &
\mathrm{ReLU}\!\Big(
V_{3,m,128} * \mathrm{ReLU}\!\big(
V_{2,m,128} *  \\
& \mathrm{ReLU}\!\big(
V_{1,m,128} * \psi_{y,m}(x,y,C)
\big)
\big)
\Big)
\Big)
\end{aligned}
\tag{6}
\end{equation}

Here, $\psi_{128}$ denotes that 128 wave functions are generated for the given object image. $H_{1,m,128}$, $H_{2,m,128}$, and $H_{3,m,128}$ represent the first, second, and third convolutional layers, each consisting of 128 filters, applied to $\psi_{x,m}$ for $m=\pm1,\pm2$. Similarly, $V_{1,m,128}$, $V_{2,m,128}$, and $V_{3,m,128}$ denote the first, second, and third convolutional layers with 128 filters applied to $\psi_{y,m}$ for $m=\pm1,\pm2$.

 \begin{figure*}[t]
\centering
\includegraphics[width=1\linewidth]{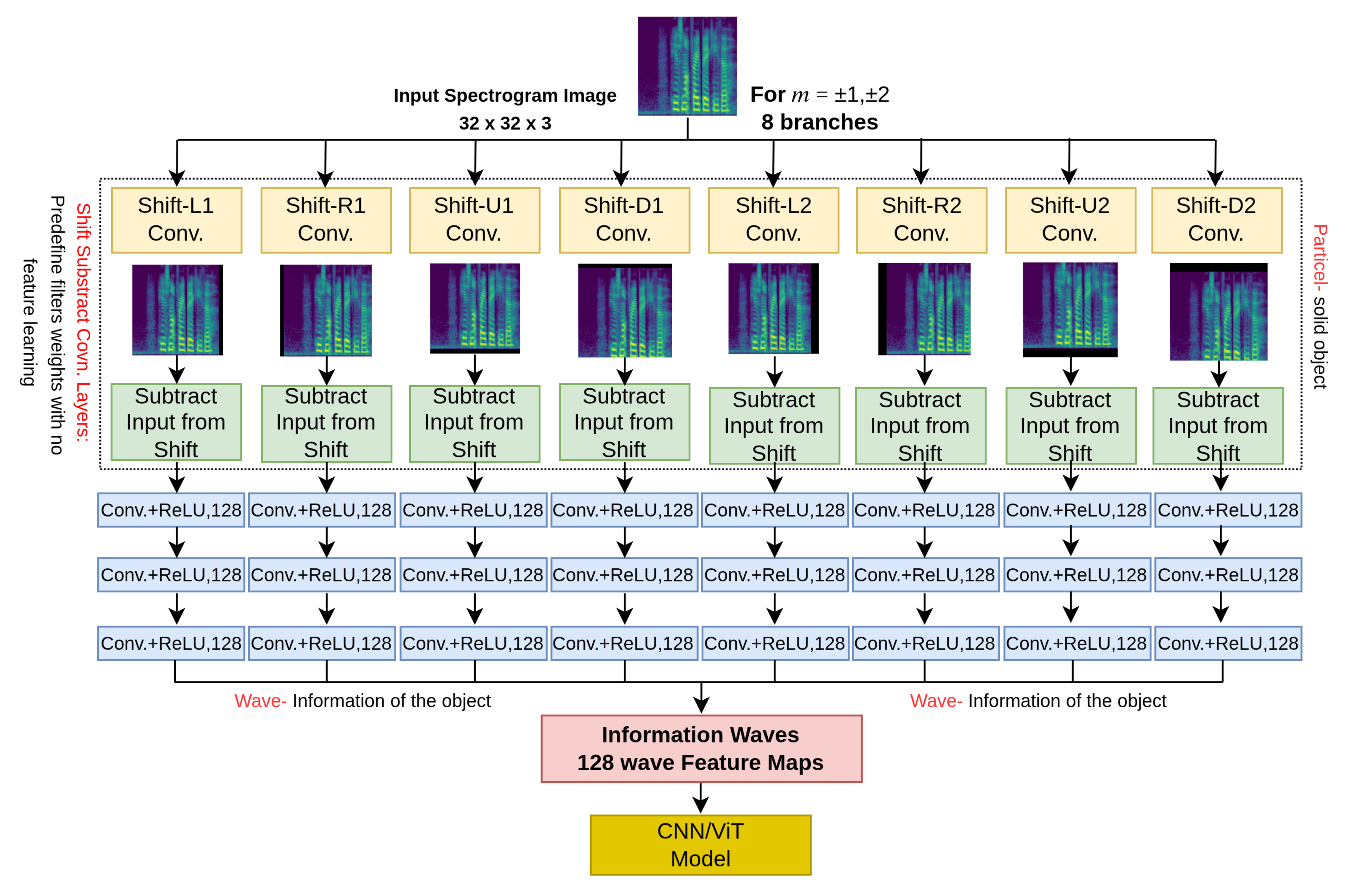} 
\caption{QV block architecture.}

\label{Fig:2}
\end{figure*}

\begin{figure*}[t]
\centering
\includegraphics[width=1\linewidth]{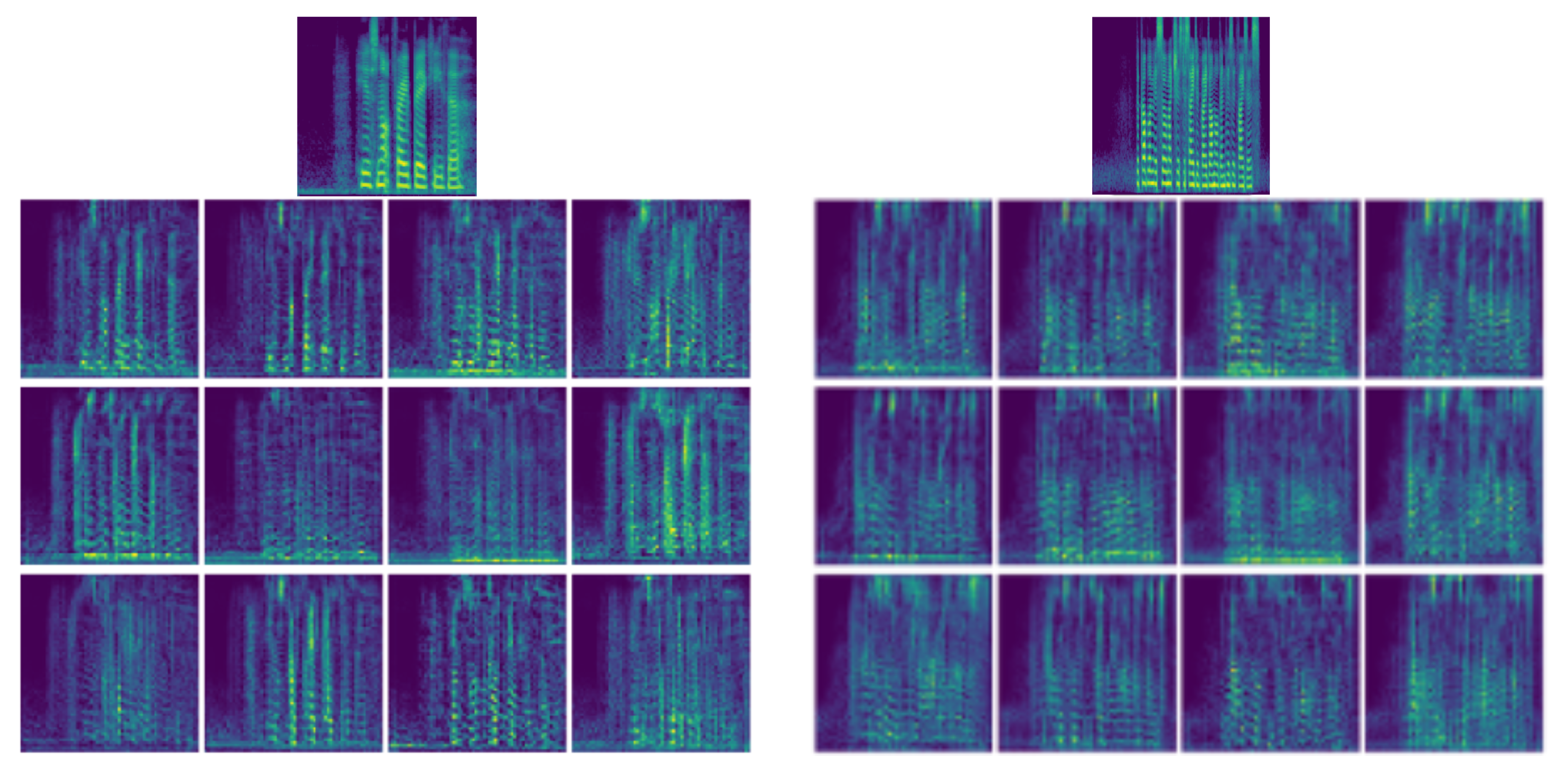}

\vspace{3pt}
\makebox[\linewidth]{%
\makebox[0.5\linewidth]{(a)}%
\makebox[0.5\linewidth]{(b)}%
}

\caption{Sample information waves. (a) Bonafide spectrogram and its corresponding information waves. (b) Spoof spectrogram and its corresponding information waves.}
\label{Fig:3}
\end{figure*}

\subsection{Implementation and Visualization of Quantum Waves on Spectrogram Images}

The implementation of the QV block is based on the method introduced by Direkoğlu et al. (2025) \cite{direkouglu2025quantum}. Therefore, in the proposed work, instead of feeding still images into deep learning models for object recognition, we extend the QV theory to the audio domain by utilizing spectrogram representations of audio signals for the classification of deepfake speech. To generate information waves, a QV block is employed, which takes a spectrogram image as input and transforms it into wave feature maps, as illustrated in Fig. \ref{Fig:2}.

The sample quantum waves of Mel spectrogram images are demonstrated in Fig. \ref{Fig:3}. The figure shows that from one spectrogram image several information wave sample images are generated by the QV Block. Each information wave extracts different features from the original spectrogram image, enriching the features for deepfake speech detection. 
These visualizations demonstrate how the proposed wave construction emphasizes boundary structures and spectro-temporal transitions, highlighting differences between genuine and manipulated speech patterns.

 \begin{figure*}[t]
\centering
\includegraphics[width=1\linewidth]{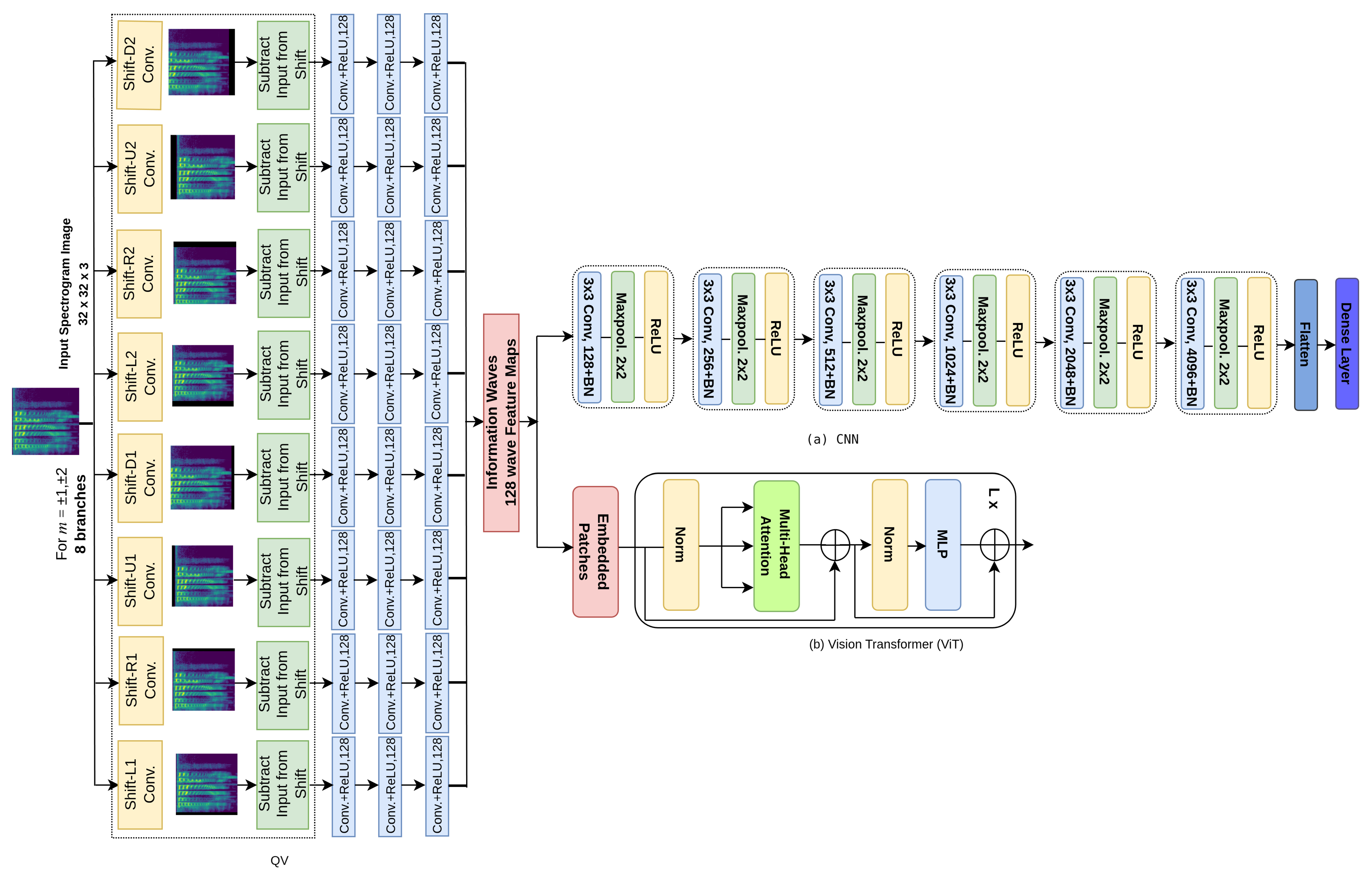} 
\caption{Workflow of QV block and deep learning of the propose study.}

\label{Fig:4}
\end{figure*}

\section {Experimental Setup and QV Model Variants}

We extend the Quantum Vision (QV) theory, originally proposed for image classification \cite{direkouglu2025quantum}, to audio classification for deepfake speech detection. Inspired by the concept of particle–wave duality, the QV framework transforms input data into information waves using a QV block before feeding them into deep learning models. This transformation enables the models to capture richer temporal–spectral patterns compared to conventional approaches. While the original work demonstrated QV using natural images such as flowers, in this study, speech spectrogram representations are used as inputs, specifically STFT spectrograms, Mel-spectrograms, and MFCCs.

All speech utterances are resampled to 16 kHz prior to feature extraction. The time–frequency representations are computed using a window length of 1024 samples and a hop length of 256 samples. For the STFT, the magnitude spectrum is obtained and converted to the decibel scale. For the Mel-spectrogram, 128 Mel filter banks are applied over the frequency range of 0–8 kHz, and the resulting power spectrum is transformed to the logarithmic scale. For MFCC, 40 coefficients are derived from the log-Mel representation using the discrete cosine transform, followed by per-coefficient normalization across time. To ensure consistent input dimensions for the classification models, each feature representation is resized to a fixed resolution of $32 \times 32$  pixels.

Two QV-based model variants are implemented in this study, namely QV-CNN and QV-ViT. In the QV-CNN variant, the QV block generates 128 wave feature maps, which are then integrated into a sequential convolutional neural network architecture referred to as QV-CNN-Heavy. This network consists of six convolutional layers, where each layer is followed by batch normalization, max-pooling, and a ReLU activation function. After the final convolutional layer, the extracted feature maps are passed through a flatten layer and then into a fully connected layer for classification. 
In the QV-ViT variant, the QV block is integrated into a Vision Transformer (ViT) framework \cite{dosovitskiy2021image}, where the 128 wave feature maps serve as input tokens. These feature maps are generated using using $m = \pm1, \pm2$, corresponding to pixel shifts of 1 and 2 in all directions. The model employs a ViT-8/8 architecture consisting of eight transformer layers with a patch size of $8 \times 8$. Each transformer layer includes layer normalization, multi-head self-attention with four attention heads, and a multi-layer perceptron block with residual connections. The MLP has a size of 2048 with a hidden dimension of 1024, enabling the model to effectively capture global contextual dependencies in the QV-transformed feature space.

The overall workflow of the proposed system, including spectrogram generation, QV transformation, and model architectures, is illustrated in Fig.~\ref{Fig:4}.

\begin{table}[t]
\begin{center}
\caption{Statistics of the ASVspoof 2019 Datasets (Durations with Three Values Denoted with Minimum/Average/Maximum).}
\label{tab:1}

\small   

\begin{tabular}{l c c c c}
\hline
\textbf{Dataset} & \textbf{Bonafide} & \textbf{Spoof} & \textbf{Total} & \textbf{Duration (sec)} \\
\hline
Training   & 2,580 & 24,072 & 26,652 & 0.65 / 3.42 / 13.19 \\
Evaluation & 7,355 & 63,882 & 71,237 & 0.47 / 3.14 / 16.55 \\
\hline
\end{tabular}

\end{center}
\end{table}

Experiments were conducted on the ASVspoof 2019 dataset \cite{todisco2019asvspoof}, following the standard training, development, and test splits to ensure consistency with prior studies. These predefined splits enable reliable evaluation and fair comparison of model performance. The distribution of the dataset across these splits is summarized in Table \ref{tab:1}.

Model performance was evaluated using multiple metrics, including accuracy, Equal Error Rate (EER), and confusion matrices, to provide a comprehensive assessment of classification effectiveness. Accuracy reflects the overall correctness of predictions, while EER captures the balance between false acceptance and false rejection rates. Additionally, confusion matrices offer detailed insights into class-wise performance.


\section {Evaluation and Results}

This section presents the results obtained using STFT spectrogram images with CNN and QV-CNN, ViT and QV-ViT models; Mel-spectrogram images with CNN and QV-CNN, ViT and QV-ViT models; and STFT-MFCC images with CNN and QV-CNN, ViT and QV-ViT models.
\subsection{STFT with QV-CNN and QV-ViT Transformers}
Tables~\ref{tab:2} and \ref{tab:3} summarize the performance of CNN-, ViT-, and their quantum-enhanced variants using STFT spectrogram features. Since both experiments rely on the same input representation, the comparison highlights the architectural impact of the proposed QV module.




\begin{table}[t]
\centering
\caption{Performance comparison of CNN and QV-CNN classifiers using STFT features with different batch sizes}
\label{tab:2}

\small  

\begin{tabular}{l l c c c c}
\hline
\textbf{Features} & \textbf{Classifier} & \textbf{Batch Size} & \textbf{Epochs} & \multicolumn{2}{c}{\textbf{Evaluation Metrics}} \\ \cline{5-6}
 &  &  &  & \textbf{Accuracy (\%)} & \textbf{EER (\%)} \\ 
\hline
STFT & CNN & 8  & 100 & 92.30 & 13.69 \\ 
STFT & CNN & 16 & 100 & 91.73 & 18.68 \\ 
STFT & CNN & 32 & 100 & 90.72 & 16.87 \\ 
STFT & CNN & 64 & 100 & 92.38 & 14.95 \\ 
STFT & QV-CNN & 8  & 100 & 91.72 & 15.35 \\ 
STFT & QV-CNN & 16 & 100 & 91.26 & 14.43 \\ 
STFT & QV-CNN & 32 & 100 & 90.88 & 16.44 \\ 
STFT & QV-CNN & 64 & 100 & \textbf{93.26} & \textbf{11.65} \\ 
\hline
\end{tabular}

\end{table}

\begin{figure}[t]
\centering

\begin{minipage}[b]{0.38\linewidth}
    \centering
    \includegraphics[width=\linewidth]{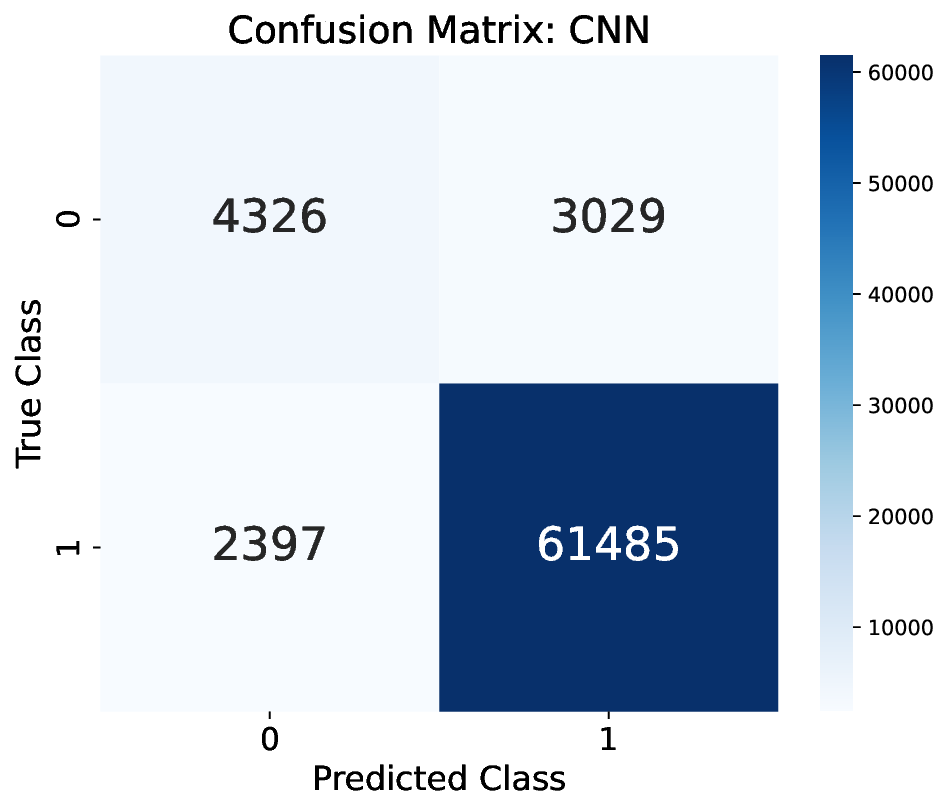}\\[2pt]
    (a) 
\end{minipage}
\hfill
\begin{minipage}[b]{0.38\linewidth}
    \centering
    \includegraphics[width=\linewidth]{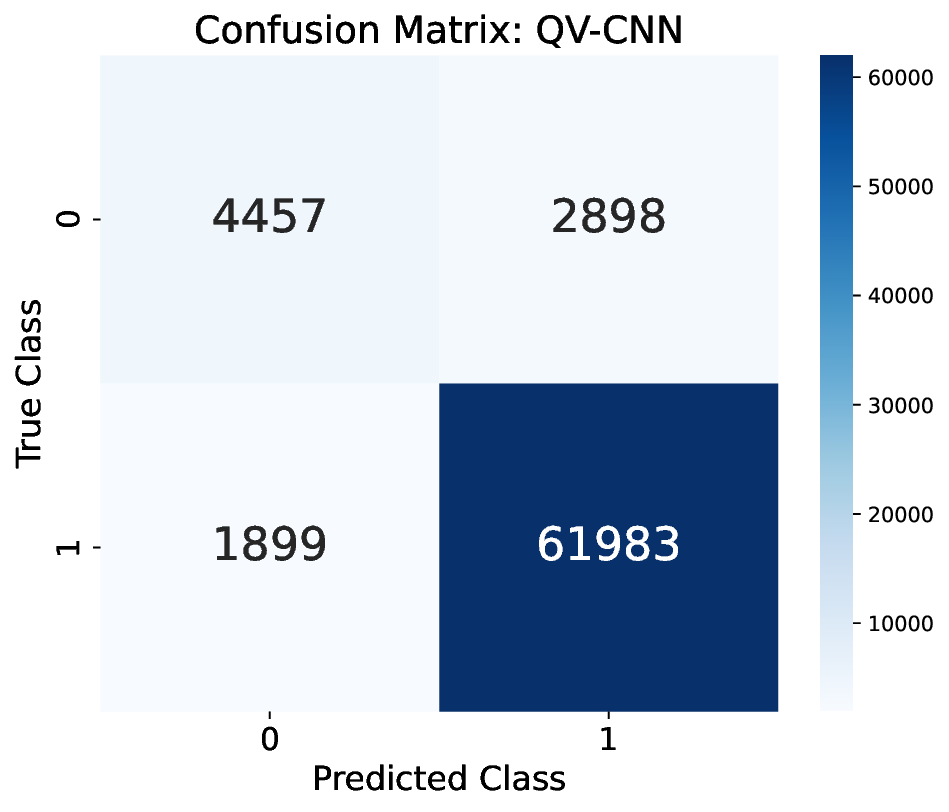}\\[2pt]
    (b)
\end{minipage}

\caption{Confusion matrices of the STFT-based (a) CNN and (b) QV-CNN evaluated on the ASVspoof evaluation set.}
\label{CM1}
\end{figure}


\begin{table}[t]
\centering
\caption{Performance comparison of ViT and QV-ViT classifiers using STFT features with different batch sizes}
\label{tab:3}

\small  

\begin{tabular}{l l c c c c}
\hline
\textbf{Features} & \textbf{Classifier} & \textbf{Batch Size} & \textbf{Epochs} & \multicolumn{2}{c}{\textbf{Evaluation Metrics}} \\ \cline{5-6}
 &  &  &  & \textbf{Accuracy (\%)} & \textbf{EER (\%)} \\ 
\hline
STFT & ViT     & 8  & 100 & 88.45 & 13.96 \\ 
STFT & ViT     & 16 & 100 & 86.34 & 12.41 \\ 
STFT & ViT     & 32 & 100 & 85.44 & 14.56 \\ 
STFT & ViT     & 64 & 100 & 85.60 & 12.91 \\ 
STFT & QV-ViT  & 8  & 100 & 89.67 & 34.14 (bias) \\ 
STFT & QV-ViT  & 16 & 100 & 89.67 & 29.15 (bias) \\ 
STFT & QV-ViT  & 32 & 100 & \textbf{90.49} & 16.28 \\ 
STFT & QV-ViT  & 64 & 100 & 89.76 & 26.92 (bias) \\ 
\hline
\end{tabular}

\end{table}

\begin{figure}[t]
\centering

\begin{minipage}[b]{0.38\linewidth}
    \centering
    \includegraphics[width=\linewidth]{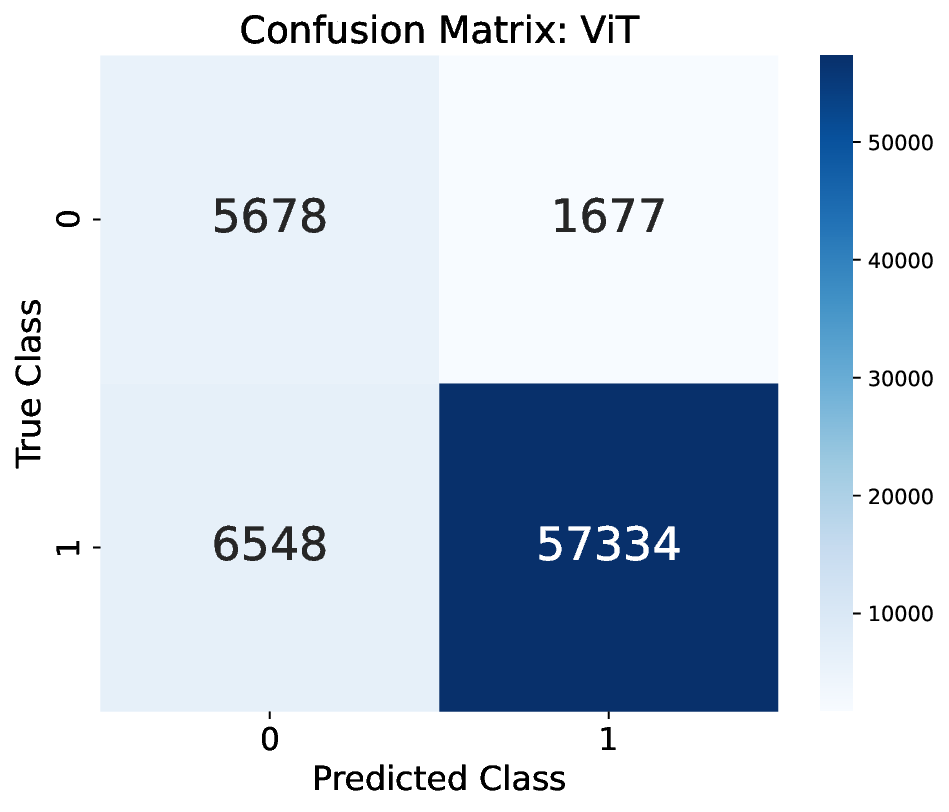}\\[2pt]
    (a) 
\end{minipage}
\hfill
\begin{minipage}[b]{0.38\linewidth}
    \centering
    \includegraphics[width=\linewidth]{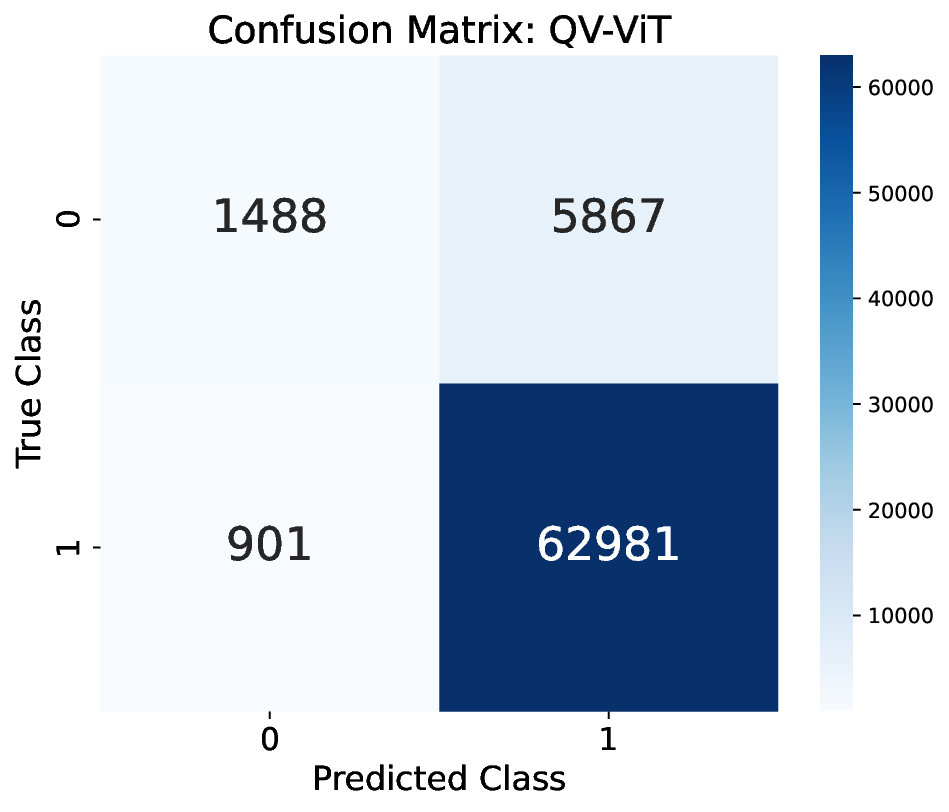}\\[2pt]
    (b)
\end{minipage}

\caption{Confusion matrices of the STFT-based (a) ViT and (b) QV-ViT evaluated on the ASVspoof evaluation set.}
\label{CM2}
\end{figure}


For CNN-based models, integrating the QV block leads to clear improvements. QV-CNN achieves the best overall performance with 93.26\% accuracy and 11.65\% EER at batch size 64, outperforming the conventional CNN across most configurations. The consistent reduction in EER indicates enhanced discriminative capability and improved robustness against deepfake artifacts.

For transformer-based models, QV-ViT improves accuracy compared to the standard ViT, reaching up to 90.49\%. However, its EER values show greater variability across batch sizes, suggesting higher sensitivity to training dynamics. 

Figures \ref{CM1}(a) and \ref{CM1}(b) show the confusion matrices corresponding to the highest accuracies achieved using STFT with CNN and STFT with QV-CNN, respectively. Similarly, Figures \ref{CM2}(a) and \ref{CM2}(b) present the confusion matrices for the best-performing models using STFT with ViT and STFT with QV-ViT, respectively.

Overall, the results demonstrate that the proposed QV module effectively enhances STFT-based representations, with particularly strong and stable improvements when integrated into CNN architectures.


\begin{table}[t]
\centering
\caption{Performance comparison of CNN and QV-CNN classifiers using Mel spectrogram features with different batch sizes}
\label{tab:4}

\small 

\begin{tabular}{l l c c c c}
\hline
\textbf{Features} & \textbf{Classifier} & \textbf{Batch Size} & \textbf{Epochs} & \multicolumn{2}{c}{\textbf{Evaluation Metrics}} \\ \cline{5-6}
 &  &  &  & \textbf{Accuracy (\%)} & \textbf{EER (\%)} \\ 
\hline
Mel & CNN    & 8  & 100 & 92.67 & 11.52 \\ 
Mel & CNN    & 16 & 100 & 92.87 & 10.71 \\ 
Mel & CNN    & 32 & 100 & 92.27 & 9.00 \\ 
Mel & CNN    & 64 & 100 & 92.23 & 14.92 \\ 
Mel & QV-CNN & 8  & 100 & 92.11 & 10.74 \\ 
Mel & QV-CNN & 16 & 100 & 92.87 & 10.71 \\ 
Mel & QV-CNN & 32 & 100 & 93.60 & 10.22 \\ 
Mel & QV-CNN & 64 & 100 & \textbf{94.57} & 10.84 \\ 
\hline
\end{tabular}

\end{table}

\begin{figure}[t]
\centering

\begin{minipage}[b]{0.38\linewidth}
    \centering
    \includegraphics[width=\linewidth]{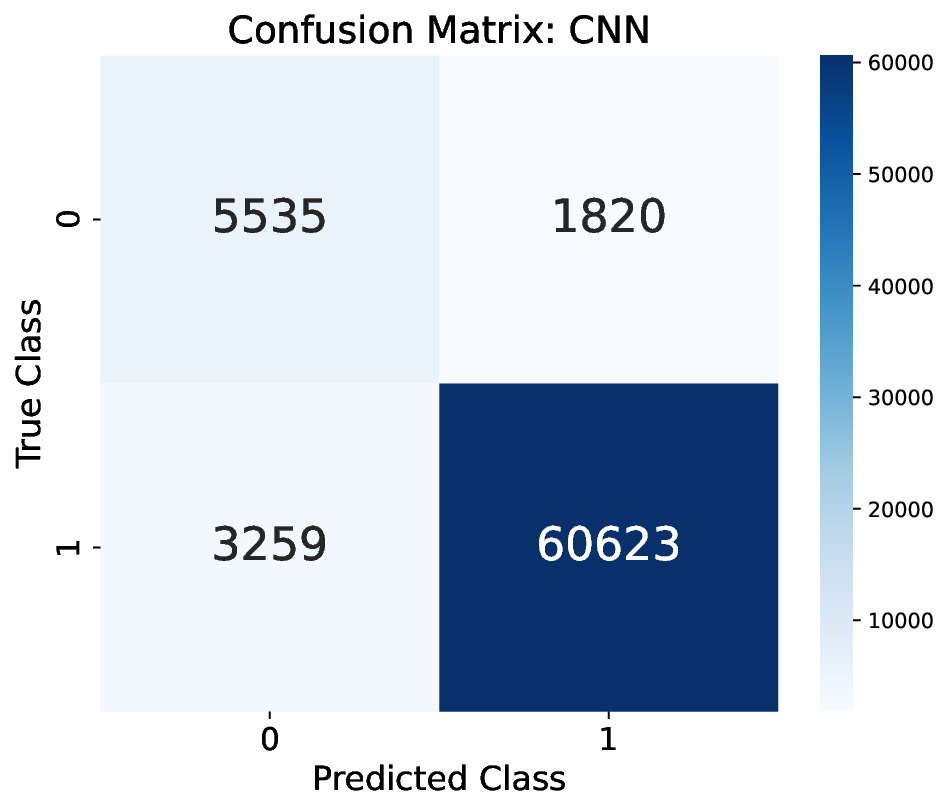}\\[2pt]
    (a) 
\end{minipage}
\hfill
\begin{minipage}[b]{0.38\linewidth}
    \centering
    \includegraphics[width=\linewidth]{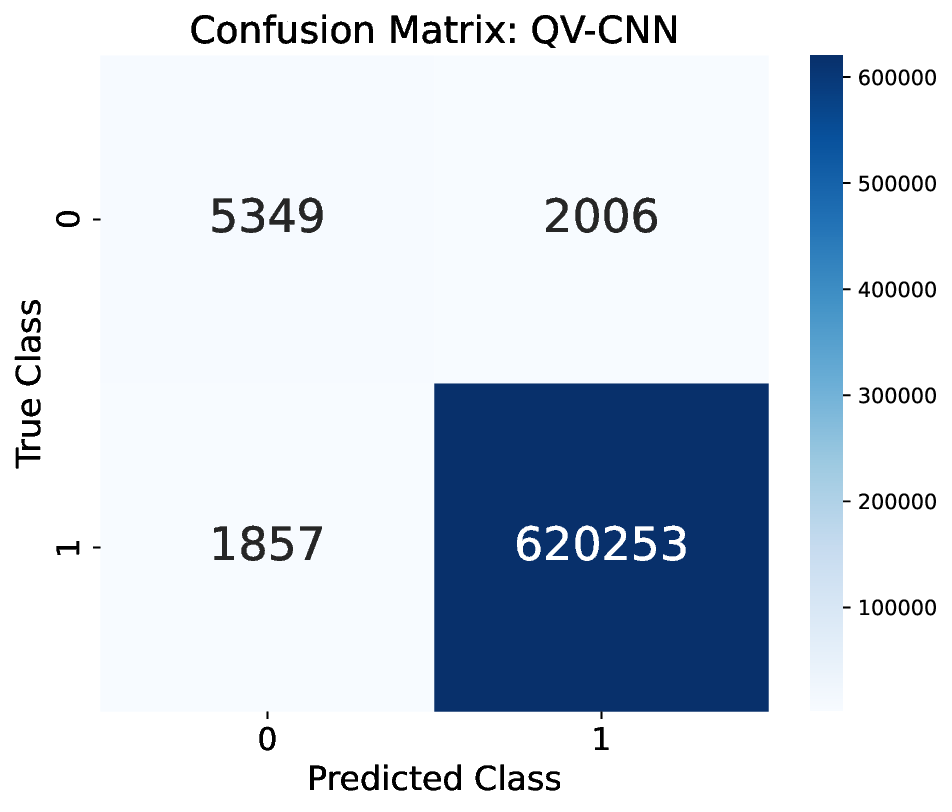}\\[2pt]
    (b)
\end{minipage}

\caption{Confusion matrices of the Mel-based (a) CNN and (b) QV-CNN evaluated on the ASVspoof evaluation set.}
\label{CM3}
\end{figure}


\begin{table}[t]
\centering
\caption{Performance comparison of ViT and QV-ViT classifiers using Mel spectrogram features with different batch sizes}
\label{tab:5}

\small 

\begin{tabular}{l l c c c c}
\hline
\textbf{Features} & \textbf{Classifier} & \textbf{Batch Size} & \textbf{Epochs} & \multicolumn{2}{c}{\textbf{Evaluation Metrics}} \\ \cline{5-6}
 &  &  &  & \textbf{Accuracy (\%)} & \textbf{EER (\%)} \\ 
\hline
Mel & ViT     & 8  & 100 & 89.65 & 18.73 \\ 
Mel & ViT     & 16 & 100 & 90.26 & 13.24 \\ 
Mel & ViT     & 32 & 100 & --    & --    \\ 
Mel & ViT     & 64 & 100 & 88.29 & 14.14 \\ 
Mel & QV-ViT  & 8  & 100 & 92.62 & 9.80 \\ 
Mel & QV-ViT  & 16 & 100 & 92.54 & 11.48 \\ 
Mel & QV-ViT  & 32 & 100 & 93.07 & 12.37 \\ 
Mel & QV-ViT  & 64 & 100 & \textbf{93.36} & 10.48 \\ 
\hline
\end{tabular}

\end{table}

\begin{figure}[t]
\centering

\begin{minipage}[b]{0.38\linewidth}
    \centering
    \includegraphics[width=\linewidth]{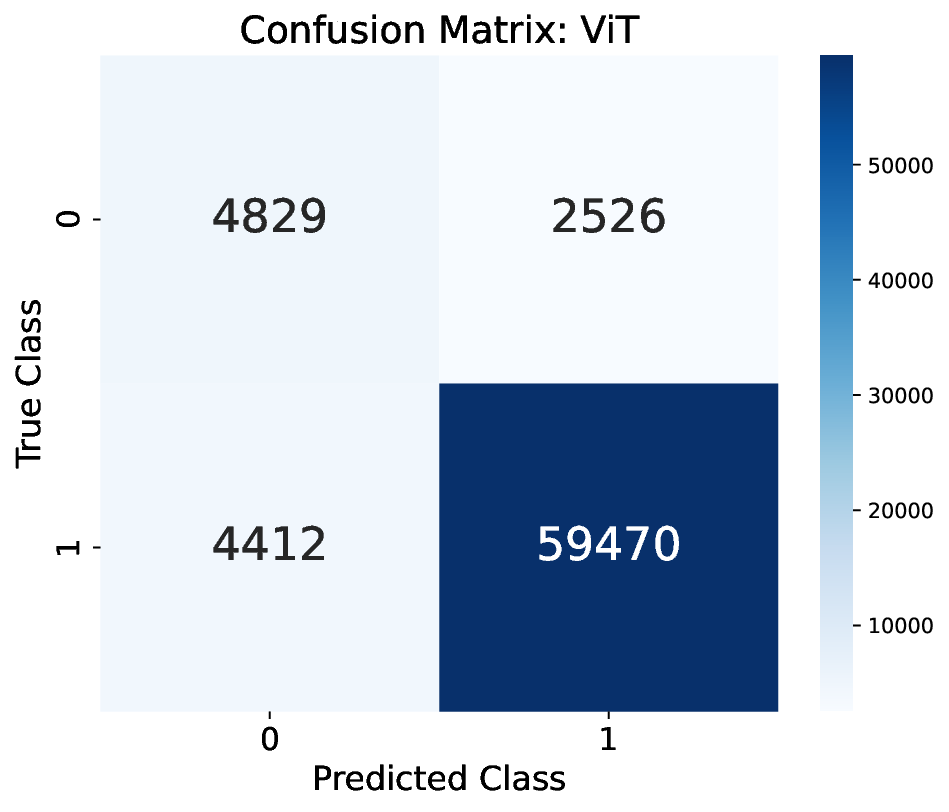}\\[2pt]
    (a) 
\end{minipage}
\hfill
\begin{minipage}[b]{0.38\linewidth}
    \centering
    \includegraphics[width=\linewidth]{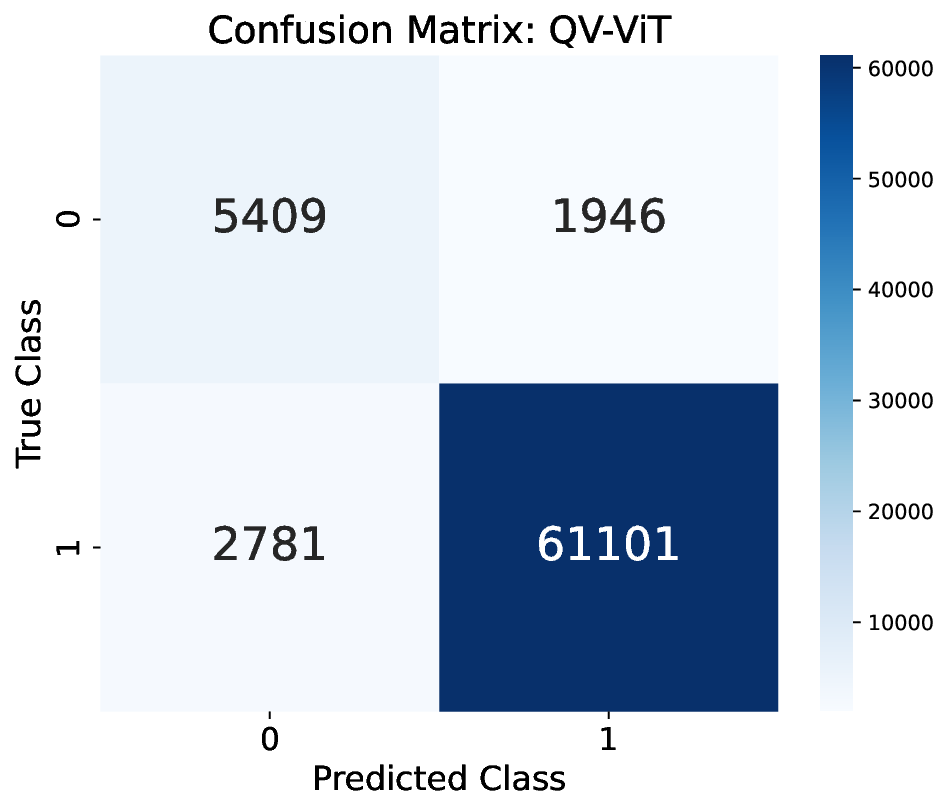}\\[2pt]
    (b)
\end{minipage}

\caption{Confusion matrices of the Mel-based (a) ViT and (b) QV-ViT evaluated on the ASVspoof evaluation set.}
\label{CM4}
\end{figure}



\begin{table}[t]
\centering
\caption{Performance comparison of CNN and QV-CNN classifiers using MFCC features with different batch sizes}
\label{tab:6}

\small 

\begin{tabular}{l l c c c c}
\hline
\textbf{Features} & \textbf{Classifier} & \textbf{Batch Size} & \textbf{Epochs} & \multicolumn{2}{c}{\textbf{Evaluation Metrics}} \\ \cline{5-6}
 &  &  &  & \textbf{Accuracy (\%)} & \textbf{EER (\%)} \\ 
\hline
MFCC & CNN    & 8  & 100 & 91.72 & 15.35 \\ 
MFCC & CNN    & 16 & 100 & 91.52 & 16.45 \\ 
MFCC & CNN    & 32 & 100 & 91.51 & 14.85 \\ 
MFCC & CNN    & 64 & 100 & 91.67 & 14.62 \\ 
MFCC & QV-CNN & 8  & 100 & 92.63 & 9.80 \\ 
MFCC & QV-CNN & 16 & 100 & 93.78 & 10.44 \\ 
MFCC & QV-CNN & 32 & 100 & 93.33 & 12.47 \\ 
MFCC & QV-CNN & 64 & 100 & \textbf{94.20} & \textbf{9.04} \\ 
\hline
\end{tabular}

\end{table}

\begin{figure}[t]
\centering

\begin{minipage}[b]{0.38\linewidth}
    \centering
    \includegraphics[width=\linewidth]{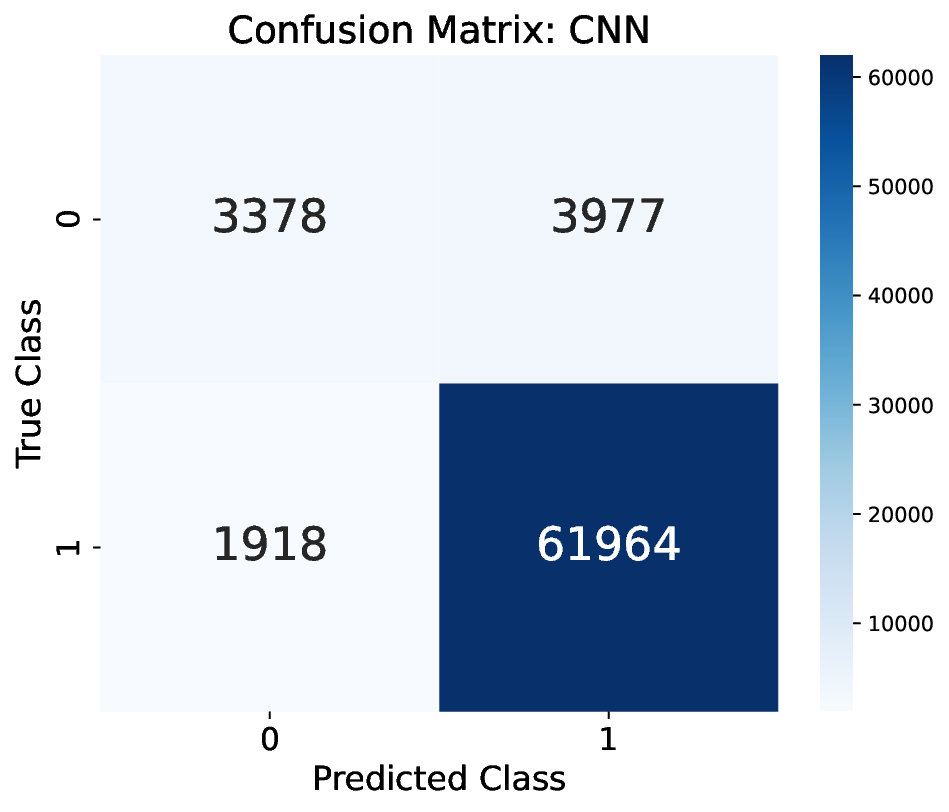}\\[2pt]
    (a) 
\end{minipage}
\hfill
\begin{minipage}[b]{0.38\linewidth}
    \centering
    \includegraphics[width=\linewidth]{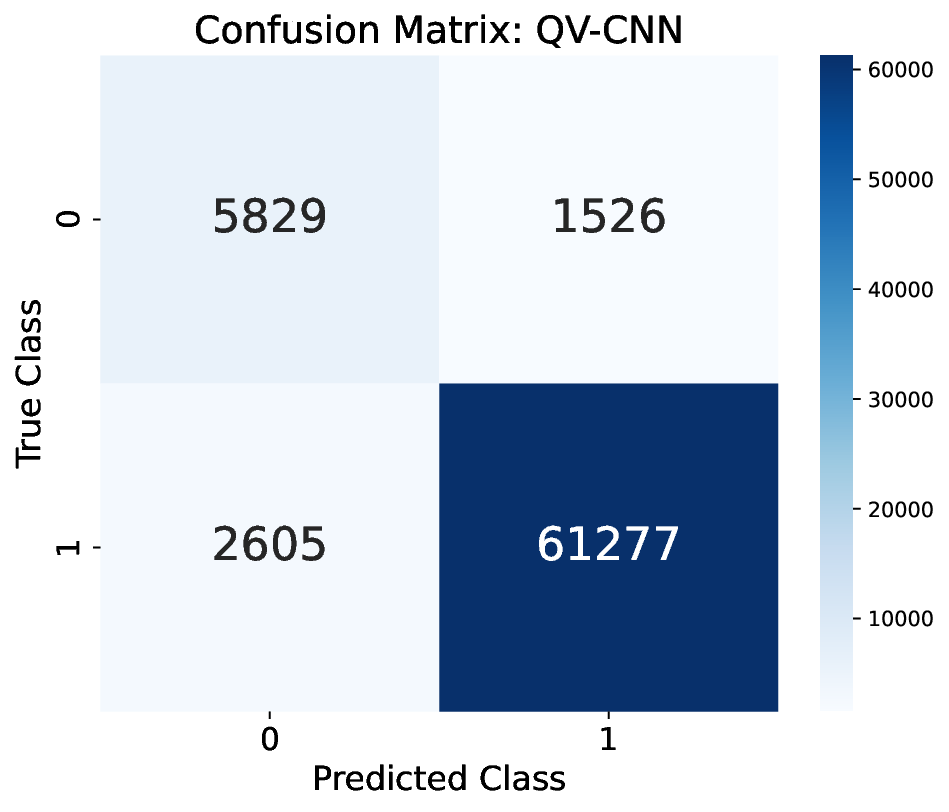}\\[2pt]
    (b)
\end{minipage}

\caption{Confusion matrices of the MFCC-based (a) CNN and (b) QV-CNN evaluated on the ASVspoof evaluation set.}
\label{CM5}
\end{figure}


\begin{table}[t]
\centering
\caption{Performance comparison of ViT and QV-ViT classifiers using MFCC features with different batch sizes}
\label{tab:7}

\small 

\begin{tabular}{l l c c c c}
\hline
\textbf{Features} & \textbf{Classifier} & \textbf{Batch Size} & \textbf{Epochs} & \multicolumn{2}{c}{\textbf{Evaluation Metrics}} \\ \cline{5-6}
 &  &  &  & \textbf{Accuracy (\%)} & \textbf{EER (\%)} \\ 
\hline
MFCC & ViT     & 8  & 100 & 89.67 & 24.39 \\ 
MFCC & ViT     & 16 & 100 & 89.06 & 20.10 \\ 
MFCC & ViT     & 32 & 100 & 89.15 & 15.67 \\ 
MFCC & ViT     & 64 & 100 & 87.07 & 17.59 \\ 
MFCC & QV-ViT  & 8  & 100 & \textbf{93.49} & 10.60 \\ 
MFCC & QV-ViT  & 16 & 100 & 91.04 & 14.50 \\ 
MFCC & QV-ViT  & 32 & 100 & 92.24 & \textbf{9.76} \\ 
MFCC & QV-ViT  & 64 & 100 & 91.07 & 11.65 \\ 
\hline
\end{tabular}

\end{table}

\begin{figure}[t]
\centering

\begin{minipage}[b]{0.38\linewidth}
    \centering
    \includegraphics[width=\linewidth]{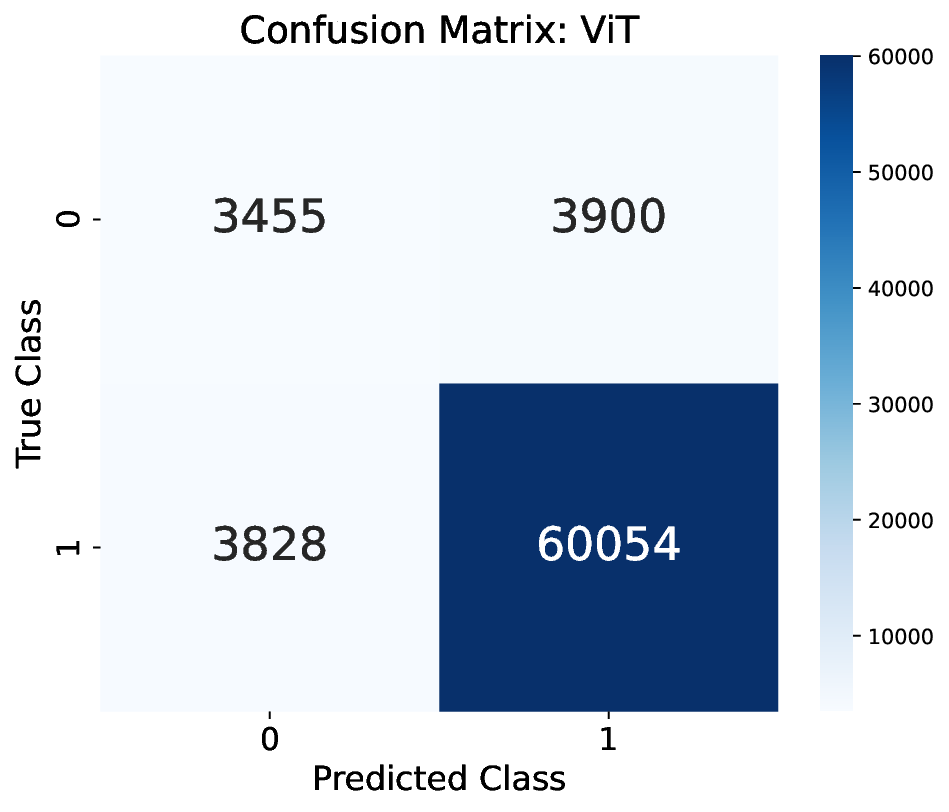}\\[2pt]
    (a) 
\end{minipage}
\hfill
\begin{minipage}[b]{0.38\linewidth}
    \centering
    \includegraphics[width=\linewidth]{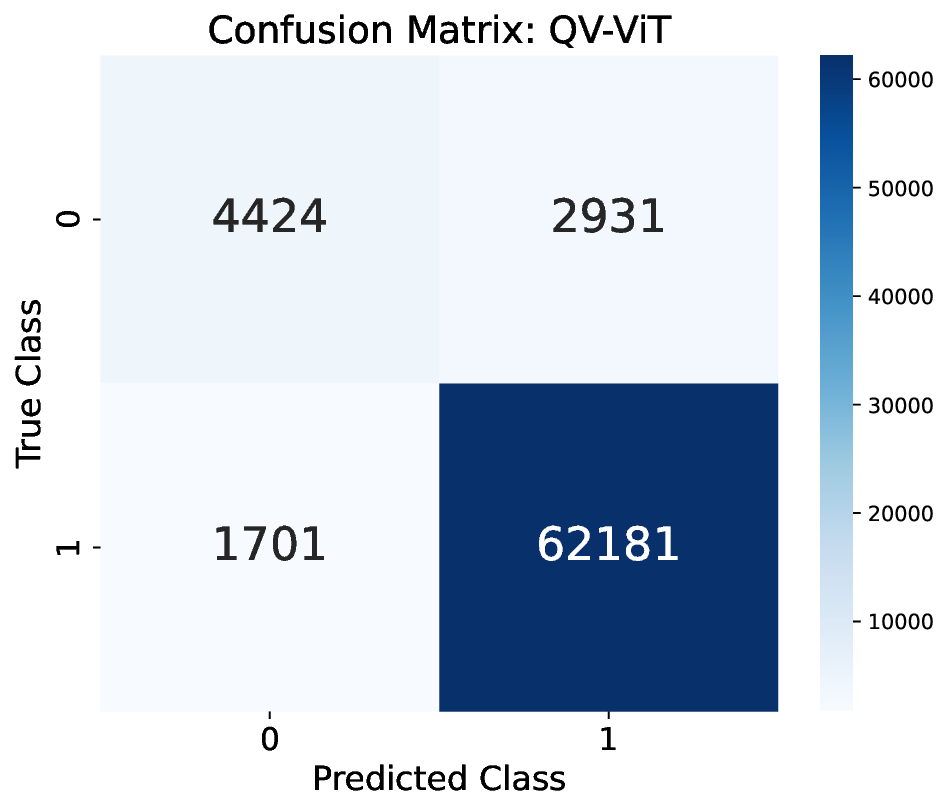}\\[2pt]
    (b)
\end{minipage}

\caption{Confusion matrices of the MFCC-based (a) CNN and (b) QV-CNN evaluated on the ASVspoof evaluation set.}
\label{CM6}
\end{figure}

\subsection{Mel Spectrogram with QV-CNN and QV-ViT Transformers}
Tables~\ref{tab:4} and \ref{tab:5} present the performance of CNN-, ViT-, and their quantum-enhanced variants using Mel spectrogram features. Compared to STFT and MFCC representations, Mel spectrograms demonstrate strong and stable classification performance across architectures.

For CNN-based models, the QV-CNN consistently improves accuracy and maintains competitive EER values. The best performance is achieved at batch size 64, reaching 94.57\% accuracy, which is the highest among all Mel-based CNN configurations. This indicates that the quantum-inspired wave transformation effectively enhances discriminative spectro-temporal patterns within Mel representations. 

For transformer-based models, the improvement is more pronounced. While the standard ViT shows moderate performance, QV-ViT significantly boosts both accuracy and EER, achieving up to 93.36\% accuracy and reducing EER to as low as 9.80\%.

Figures \ref{CM3}(a) and \ref{CM3}(b) illustrate the confusion matrices corresponding to the highest accuracies achieved using Mel-spectrogram with CNN and Mel-spectrogram with QV-CNN, respectively. Similarly, Figures \ref{CM4}(a) and \ref{CM4}(b) present the confusion matrices for the best-performing models using Mel-spectrogram with ViT and Mel-spectrogram with QV-ViT, respectively.

Overall, the Mel-spectrogram-based experiments further confirm the effectiveness of the proposed QV module, particularly in reducing error rates and improving classification stability across different architectures

\subsection{MFCC with QV-CNN and QV-ViT Transformers}
Tables~\ref{tab:6} and \ref{tab:7} present the performance of CNN-, ViT-, and their quantum-enhanced variants using MFCC features. Unlike the STFT-based results, MFCC representations show more consistent gains when integrated with the proposed QV module.

For CNN architectures, QV-CNN significantly improves both accuracy and EER across all batch sizes. The best performance is achieved at batch size 64 with 94.20\% accuracy and 9.04\% EER, demonstrating substantial error reduction compared to the conventional CNN. This indicates that the quantum-inspired wave transformation effectively enhances discriminative information within MFCC representations.

For transformer-based models, the improvement is even more pronounced. While the standard ViT exhibits relatively high EER values, QV-ViT achieves marked performance gains, reaching 93.49\% accuracy and as low as 9.76\% EER. The consistent reduction in EER suggests that the QV representation strengthens spectro-temporal feature encoding, making the model more robust to deepfake artifacts. 

Figures \ref{CM5}(a) and \ref{CM5}(b) show the confusion matrices corresponding to the highest accuracies achieved using Mel-spectrogram with CNN and MFCC with QV-CNN, respectively. Similarly, Figures \ref{CM6}(a) and \ref{CM6}(b) present the confusion matrices for the best-performing models using MFCC with ViT and Mel-spectrogram with QV-ViT, respectively.

Overall, the MFCC-based experiments further confirm the effectiveness of the proposed QV module, particularly in reducing error rates and improving classification stability across different architectures.


 \begin{figure*}[t]
\centering

\begin{minipage}[b]{0.40\linewidth}
    \centering
    \includegraphics[width=\linewidth]{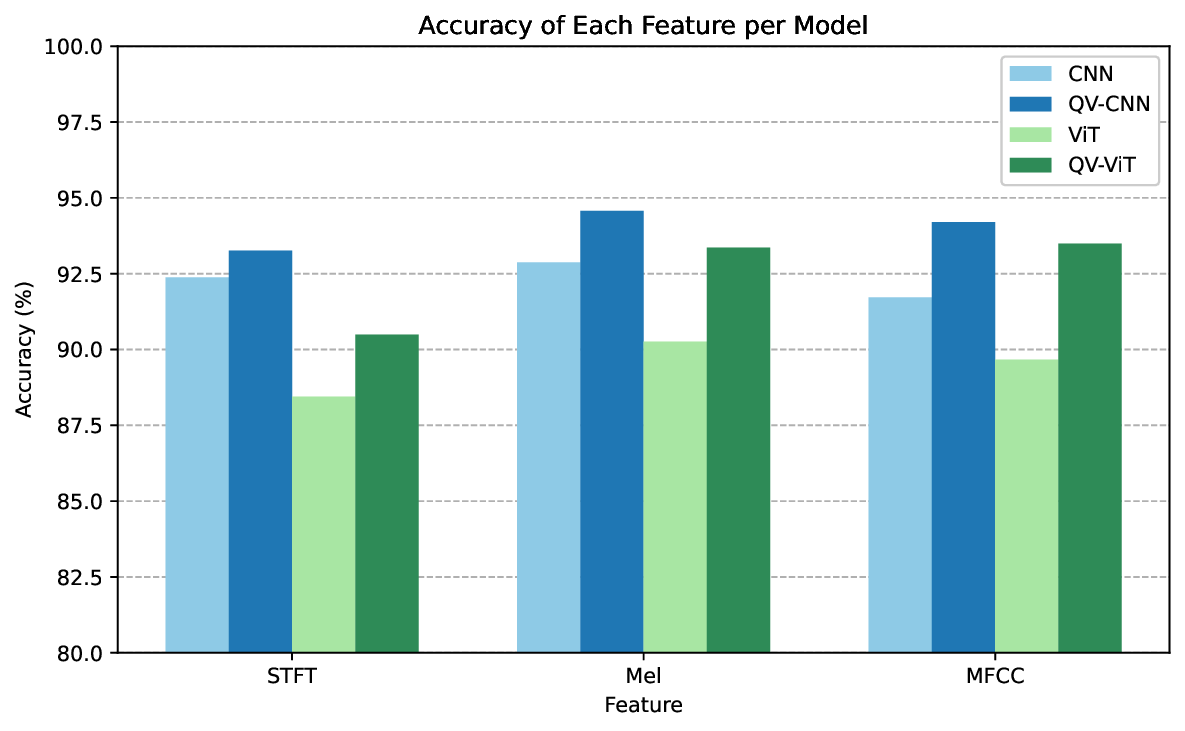}\\[2pt]
    (a) 
\end{minipage}
\hfill
\begin{minipage}[b]{0.40\linewidth}
    \centering
    \includegraphics[width=\linewidth]{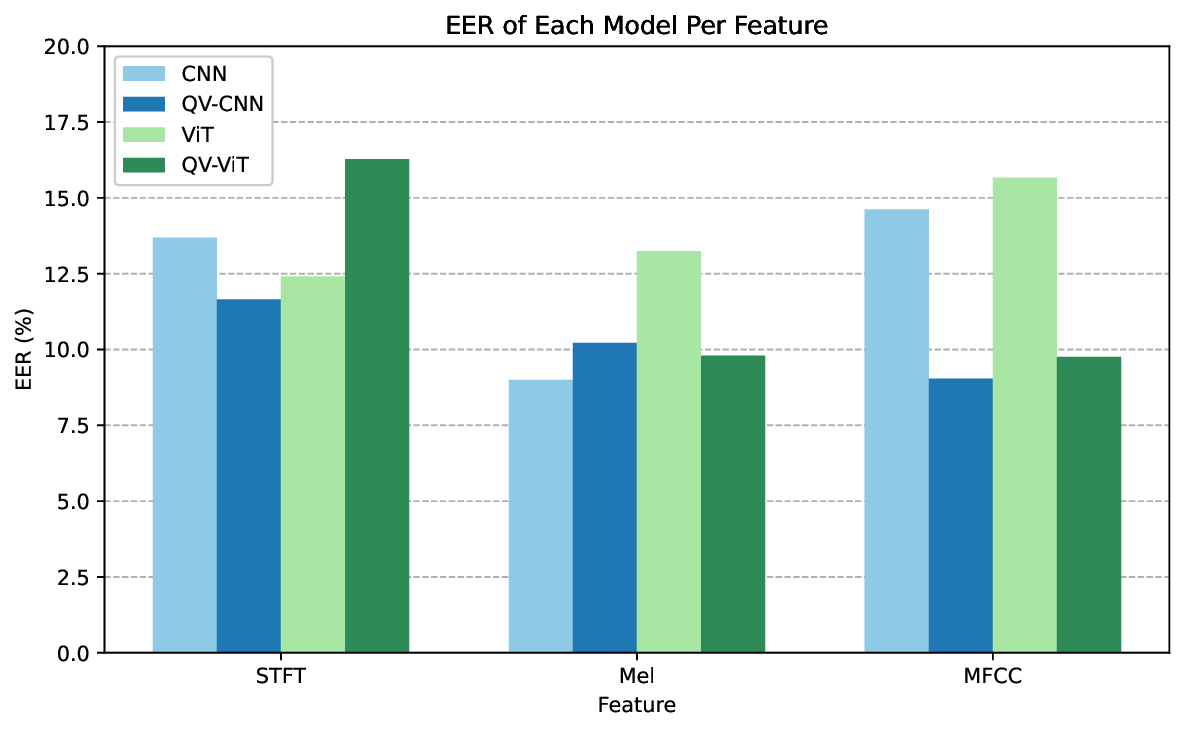}\\[2pt]
    (b)
\end{minipage} 

\caption{Accuracy (a) and ERR (b) for each model  per feature}
\label{f9}
\end{figure*}

 \begin{figure*}[t]
\centering

\begin{minipage}[b]{0.40\linewidth}
    \centering
    \includegraphics[width=\linewidth]{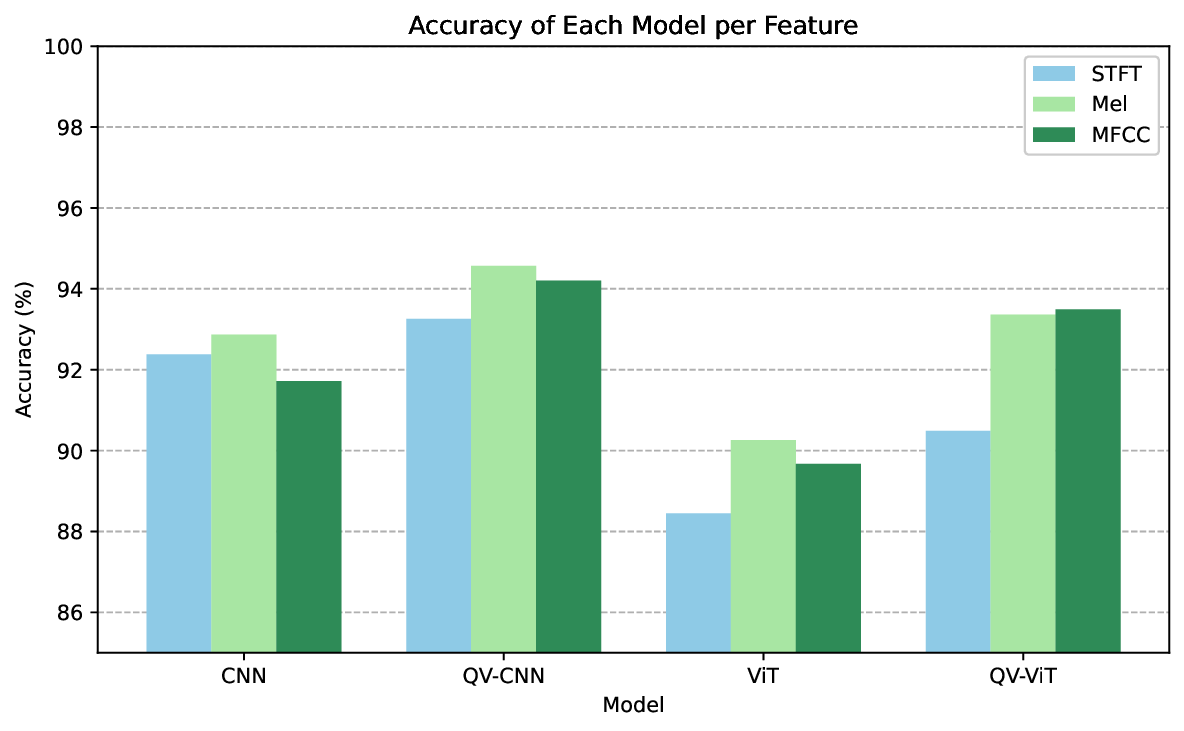}\\[2pt]
      (a) 
\end{minipage}
\hfill
\begin{minipage}[b]{0.40\linewidth}
    \centering
    \includegraphics[width=\linewidth]{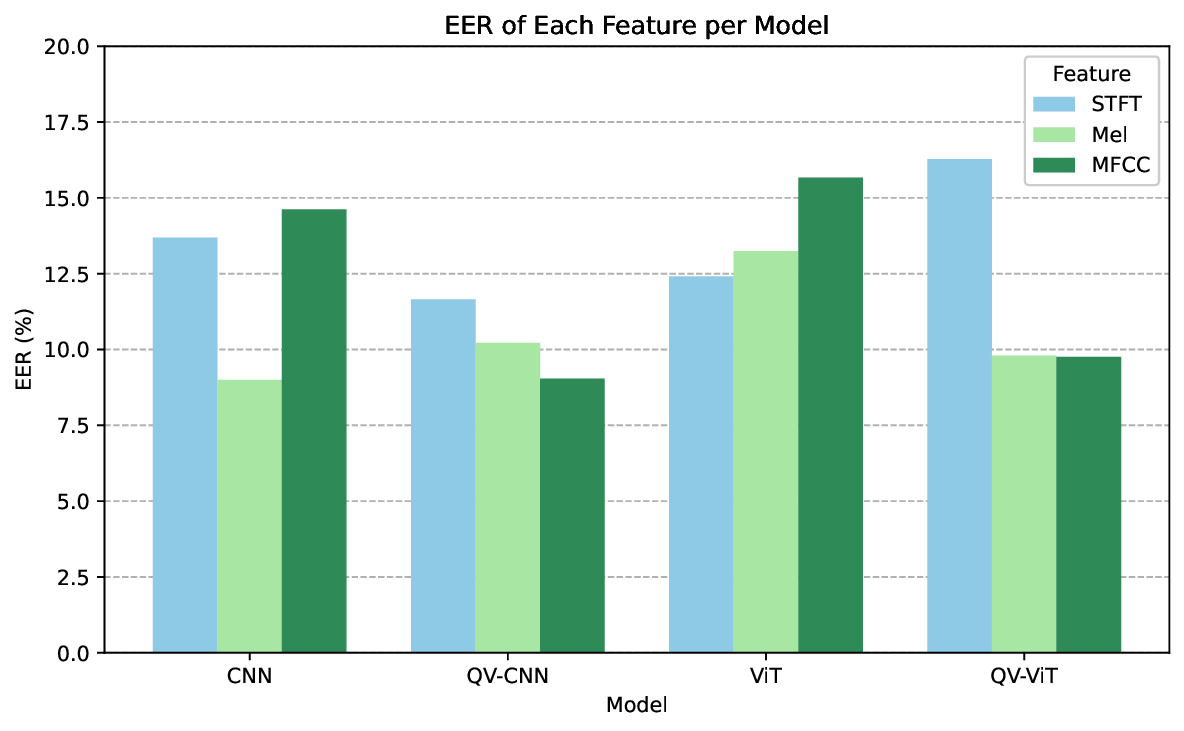}\\[2pt]
    (b)
\end{minipage}

\caption{Accuracy a) and ERR b) of each feature per model.}
\label{f10}
\end{figure*}

 \begin{figure*}[h]
\centering
\includegraphics[width=1\linewidth]{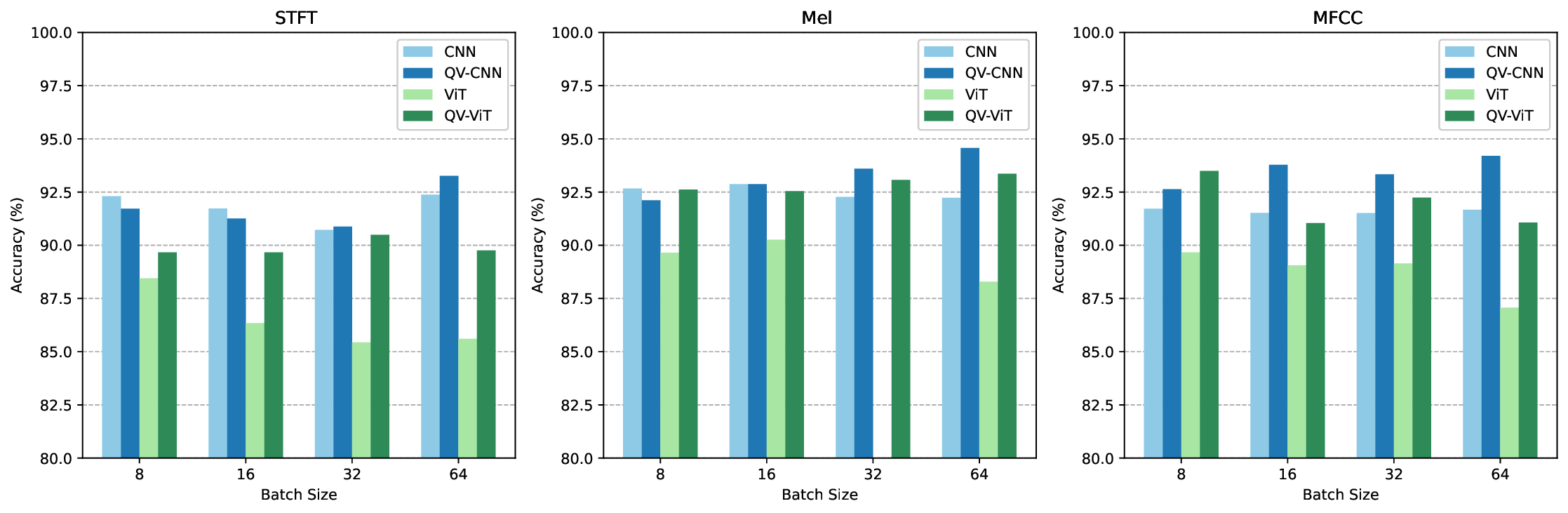} 
\caption{Accuracy  of each feature per model for batch size .}

\label{F11}
\end{figure*}

 \begin{figure*}[!h]
\centering
\includegraphics[width=1\linewidth]{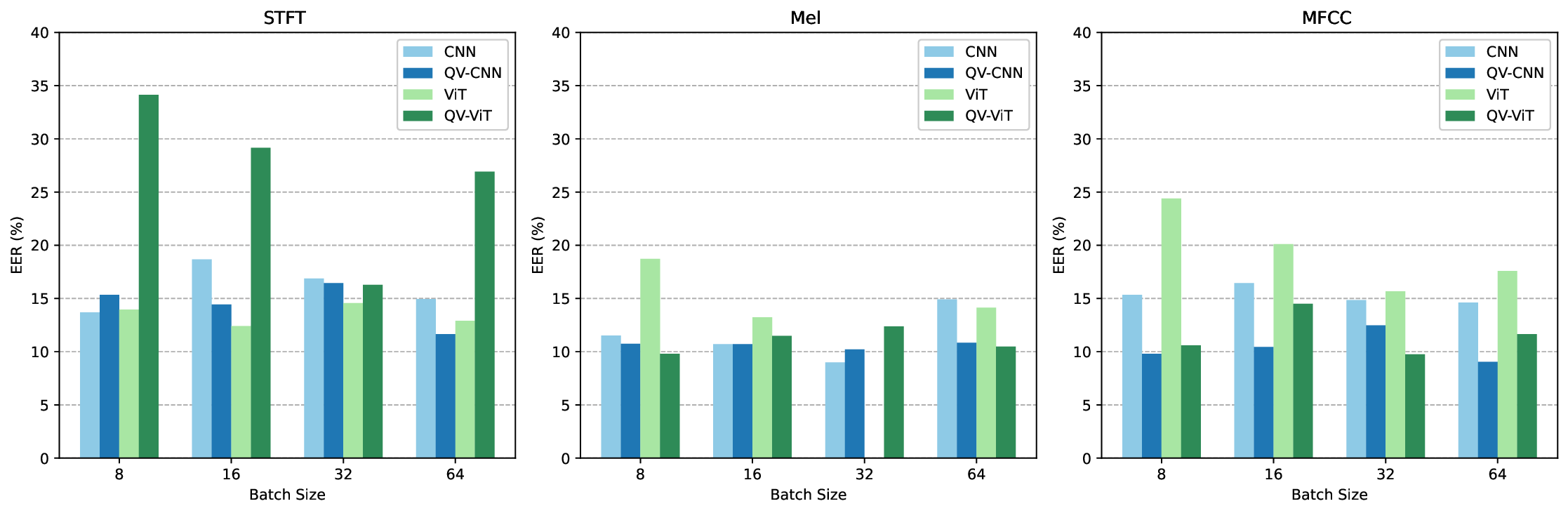} 
\caption{ ERR of each feature per model for batch size}

\label{F12}
\end{figure*}

\section{Discussion}

Figures \ref{f9} and \ref{f10} present two complementary analyses of the proposed approach. Figure \ref{f9} provides a model-centric comparison by evaluating the classification accuracy and Equal Error Rate (EER) of different architectures across three feature representations: STFT, Mel-spectrogram, and MFCC. The results show that incorporating the proposed QV mechanism consistently improves model performance. For the CNN-based architecture, the baseline CNN achieves accuracies of 92.38\%, 92.87\%, and 91.72\% for STFT, Mel, and MFCC features, respectively. After integrating the QV module, the QV-CNN model improves the performance to 93.26\%, 94.57\%, and 94.20\%. The largest improvement occurs for MFCC features, demonstrating that the QV mechanism enhances the representation capability of cepstral features within convolutional architectures.

Figure \ref{f10} presents a feature-centric analysis, where the performance of different feature representations is compared across the evaluated models. The results indicate that Mel and MFCC features provide more discriminative information than STFT for spoof detection tasks. In particular, the QV-CNN model achieves the highest accuracy for both Mel (94.57\%) and MFCC (94.20\%) features. Similarly, the transformer-based architecture benefits from the proposed QV mechanism, as the QV-ViT model improves the accuracy of the baseline ViT from 88.45\% to 90.49\% for STFT, from 90.26\% to 93.36\% for Mel, and from 89.67\% to 93.49\% for MFCC. The EER results further support these findings, with the lowest error rate of 9.04\% achieved by the QV-CNN model using MFCC features.

Figures \ref{F11} and \ref{F12} further investigate the effect of batch size on model performance. The results demonstrate that the proposed QV-based architectures maintain stable performance across different batch sizes while generally benefiting from larger batches during training. In most cases, batch size 64 achieves the highest classification accuracy and the lowest EER values. Among all configurations, the QV-CNN model combined with MFCC features consistently provides the best spoof detection performance, confirming the effectiveness of integrating the QV mechanism with cepstral feature representations.

Table \ref{tab:8} presents a performance comparison between the proposed QV-CNN models and several existing deepfake and spoofed speech detection approaches. Prior studies report EER values ranging from 9.33\% to 17.51\%, while reported accuracies vary between 85.99\% and 93.36\%. The proposed methods achieve superior performance, where QV-CNN (MFCC) attains the lowest EER of 9.04\% with an accuracy of 94.20\%, and QV-CNN (Mel spectrogram) achieves the highest accuracy of 94.57\% with a competitive EER of 10.84\%. Compared to traditional spectral and cepstral feature-based methods (e.g., CQCC, MFCC, STM) and recent transformer-based architectures, the proposed QV-CNN framework provides improved discriminative capability while maintaining robustness across evaluation metrics. These results indicate that the proposed approach effectively enhances spoof detection performance relative to existing methods.

\begin{table}[!t]
\centering
\caption{Performance comparison with existing studies}
\label{tab:8}

\small 

\begin{tabular}{p{2cm} p{6.8cm} c c}
\hline
\textbf{Study} & \textbf{Approach} & \multicolumn{2}{c}{\textbf{Evaluation Metrics}} \\
\cline{3-4}
 &  & \textbf{Accuracy (\%)} & \textbf{EER (\%)} \\
\hline

\cite{altalahin2023unmasking} & MFCC + LSTM-CNN & 88.00 & -- \\
\cite{cheng2023analysis} & STM (Mel FB) + LCNN & -- & 9.79 \\
\cite{ulutas2023deepfake} & Spectrogram + Vision Transformer & -- & 11.02 \\ 
\cite{bartusiak2021synthesized} & Compact Convolutional Transformer (CCT) & 92.13 & -- \\ 
\cite{das2020assessing} & CQCC + DNN & -- & 13.74 \\
\cite{nosek2019synthesized} & Spectrogram + CNN & -- & 9.57 \\
\cite{alzantot2019deep} & MFCC + ResNet & -- & 9.33 \\
\cite{bartusiak2021frequency} & Spectrogram + CNN & 85.99 & -- \\
\cite{wang2020asvspoof} & CQCC + GMM (B1) & -- & 9.57 \\ 
\cite{talagini2024audio} & Benford Distribution + Transformer & -- & 16.37 \\ 
\cite{cuccovillo2023audio} & Log-Spectrogram + Formant Transformer & -- & 17.51 \\ 
\cite{zaman2024hybrid} & STFT + VGG-Transformer & 93.36 & -- \\ 

\hline
\textbf{Our} & \textbf{QV-CNN (MFCC)} & \textbf{94.20} & \textbf{9.04} \\
\textbf{Our} & \textbf{QV-CNN (Mel Spectrogram)} & \textbf{94.57} & 10.84 \\
\hline

\end{tabular}
\end{table}

\FloatBarrier
\section{Conclusion}
In this work, we introduced Quantum Vision (QV) theory as a novel perspective for deep learning–based audio classification and applied it to deepfake speech detection. Inspired by the particle–wave duality principle in quantum physics, the proposed framework transforms conventional spectrogram representations into quantum-inspired information waves before classification. This transformation is implemented through a QV block and integrated into Convolutional Neural Networks (QV-CNN) and Vision Transformers (QV-ViT), enabling end-to-end learning.

Extensive experiments were conducted using STFT, MFCC, and Mel spectrogram representations on the ASVspoof benchmark dataset. Across different feature types and model architectures, QV-based models consistently outperformed their non-QV counterparts. In particular, the QV-CNN model combined with Mel spectrogram features achieved the best overall performance, reaching 94.57\% classification accuracy and an Equal Error Rate (EER) of 9.04\%, demonstrating state-of-the-art effectiveness on the evaluated dataset. The consistent reduction in EER further indicates improved robustness in distinguishing genuine and spoofed speech signals.

The results confirm that transforming spectrograms into quantum-inspired information wave representations enhances discriminative feature learning and improves deepfake detection reliability. These findings demonstrate that QV theory provides an effective and promising direction for audio deepfake detection and opens new avenues for quantum-inspired learning in audio perception and classification tasks.

Future work will explore extending QV theory to other audio representations, larger-scale datasets, and multimodal deepfake detection frameworks.

\section*{Acknowledgment}

This paper is under consideration for publication in
\textit{Pattern Recognition Letters}.
\section*{Declaration of interests}

The authors declare that they have no competing interests.

\bibliographystyle{ieeetr}
\bibliography{refs}

\end{document}